\documentclass[review]{elsarticle}

\usepackage{lineno,hyperref}

\usepackage{multirow}
\usepackage{float}
\usepackage{booktabs}
\usepackage{xcolor}
\usepackage{url}
\usepackage{color}
\usepackage{array}
\usepackage{tikz}
\usepackage{hyperref}
\usepackage{colortbl}
\usepackage{subfigure}
\usepackage{graphicx}
\usepackage{soul}
\usepackage{amsmath}
\usepackage{amssymb}

\newcommand{\tabincell}[2]{\begin{tabular}{@{}#1@{}}#2\end{tabular}}
\definecolor{mycyan}{rgb}{0.88,0.90,0.94}

\usepackage{mathrsfs,amsmath}
\usepackage{algorithm}
\usepackage{algorithmicx}
\usepackage{algpseudocode}
\usepackage{multirow}
\usepackage{float}
\usepackage{booktabs}
    
\modulolinenumbers[5]

\journal{Journal of \LaTeX\ Templates}

\begin{document}

\begin{frontmatter}

\title{Tailored Multi-Organ Segmentation with Model Adaptation and Ensemble}


\author{Jiahua Dong$^{1}$}\author{Guohua Cheng$^{1}$}\author{Yue Zhang$^{2, 5}$\corref{mycorrespondingauthor}}\author{Chengtao Peng$^4$}\author{Yu Song$^3$}\author{Ruofeng Tong$^1$\corref{mycorrespondingauthor}}\author{Lanfen Lin$^1$}\author{Yen-Wei Chen$^3$}

\address{1 College of Computer Science and Technology, Zhejiang University, Hangzhou, 310027, China \\
\vspace{5pt}
2 Center for Medical Imaging, Robotics, Analytic Computing \& Learning (MIRACLE), Suzhou Institute for Advanced Research, University of Science and Technology of China, Suzhou, 215163, China \\
\vspace{5pt}
3 Graduate School of Information Science and Engineering, Ritsumeikan University, Shiga, 525-8577, Japan \\
\vspace{5pt}

4 Department of Electronic Engineering and Information Science, University of Science and Technology of China, Hefei, 230026, China\\
\vspace{5pt}
5 School of Biomedical Engineering, Division of Life Sciences and Medicine, University of Science and Technology of China, Hefei, Anhui, 230026, China
}









\cortext[mycorrespondingauthor]{ Corresponding author (yue\_zhang@mail.ustc.edu.cn, trf@zju.edu.cn)}

\begin{abstract}
Multi-organ segmentation, which identifies and separates different organs in medical images, is a fundamental task in medical image analysis. Recently, the immense success of deep learning motivated its wide adoption in multi-organ segmentation tasks. However, due to expensive labor costs and expertise, the availability of multi-organ annotations is usually limited and hence poses a challenge in obtaining sufficient training data for deep learning-based methods. In this paper, we aim to address this issue by combining off-the-shelf single-organ segmentation models to develop a multi-organ segmentation model on the target dataset, which helps get rid of the dependence on annotated data for multi-organ segmentation. To this end, we propose a novel dual-stage method that consists of a Model Adaptation stage and a Model Ensemble stage. The first stage enhances the generalization of each off-the-shelf segmentation model on the target domain, while the second stage distills and integrates knowledge from multiple adapted single-organ segmentation models. Extensive experiments on four abdomen datasets demonstrate that our proposed method can effectively leverage off-the-shelf single-organ segmentation models to obtain a tailored model for multi-organ segmentation with high accuracy.

\end{abstract}


\end{frontmatter}


\section{Introduction}
\label{Intro}
Multi-organ segmentation delineates important organs (e.g., the liver, spleen, and kidney) from abdominal medical images, which is essential for various clinical applications including computer-aided surgery, computer-aided diagnosis, radiation therapy, etc~\cite{wang2019abdominal}.
Recently, 
deep learning-based methods, such as U-Net~\cite{ronneberger2015u} and nnU-Net~\cite{isensee2021nnu}, have shown promising results in multi-organ segmentation, which rely heavily on large-scale annotated data with strong supervision. However, constrained by labor costs and required expertise, the limited availability of multi-organ annotations presents a challenge in acquiring sufficient data for supervised learning~\cite{li2020model}. This limitation impedes the wider adoption of deep learning-based multi-organ segmentation in clinical applications~\cite{zhou2019prior}.

\begin{figure}[t]
\centering
\includegraphics[width=0.8\textwidth]{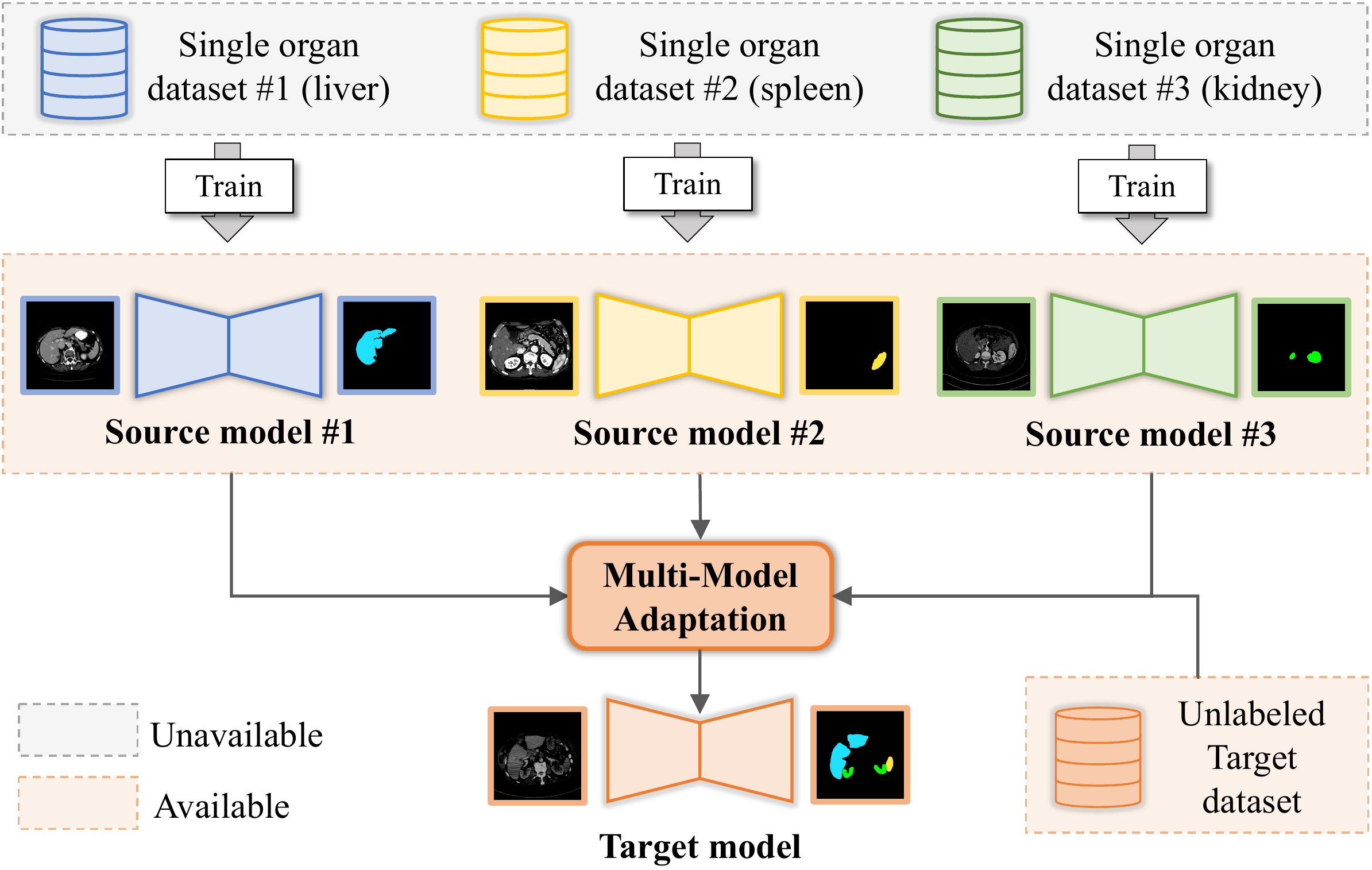}
\caption{An example of Multi-Model Adaptation, which integrates three single-organ segmentation models into a target multi-model segmentation model.}
\label{fig_exp}
\end{figure}

With the development of the open source community, many researchers released codes and well-trained medical image segmentation models to the public, but usually without providing training/testing datasets due to privacy concerns~\cite{altini2022liver, hu2016automatic, shen2019customizing}. We have observed that most of such released models are trained for single organ segmentation and retain specific knowledge distilled from their training data. Inspired by this, in this paper, we aim to explore obtaining a multi-organ segmentation model on the target dataset from a union of public single-organ segmentation models (termed as \textbf{Multi-Model Adaptation} problem), which has the potential to get rid of the dependence on annotated data for multi-organ segmentation. Fig.~\ref{fig_exp} illustrates a conceptual diagram of integrating three models, trained respectively for segmenting the liver, spleen, and kidney, into a single multi-organ segmentation model. To solve the Multi-Model Adaptation problem, there are two main challenges that need to be considered.

As public single-organ segmentation models are trained using different datasets across various institutions, 
directly applying such models to out-of-distribution target data in unseen domains usually leads to bad generalization~\cite{ganin2015unsupervised, toldo2020unsupervised, hong2022source}. Thus, the first challenge is \emph{how to adapt source segmentation models to the target domain without access to the source data}, which is also known as model adaptation.
Once we have obtained adapted single-organ models that generalize well on the target domain, the second challenge is \emph{how to aggregate multiple single-organ segmentation models into a desired multi-organ segmentation model.}

To tackle the aforementioned issues, in this paper, we propose a dual-stage method for Multi-Model Adaptation, which comprises a Model Adaptation stage and a Model Ensemble stage. 
The \textbf{Model Adaptation stage} aims to enhance the generalization of each off-the-shelf segmentation model on the target domain, without access to both source data and target labels. To achieve this, we introduce a Label Refinement Module (LRM) to produce reliable pseudo-labels for the target data and a Feature Generalization Module (FRM) to ensure the learned feature space of the target domain is robust and compact.
The \textbf{Model Ensemble stage} aims to distill and integrate knowledge from multiple adapted single-organ segmentation models. To that end, we propose a novel certainty-aware ensemble function, which can dynamically compose the teacher model from adapted single-organ models through a teacher selection map. Then, the information is distilled and transferred from the teacher model to the target multi-organ segmentation model.
To verify the performance of our framework, we conduct extensive experiments on four abdominal datasets. 
Experimental results demonstrate that our approach can effectively leverage public single-organ segmentation models to obtain a tailored model for multi-organ segmentation with high accuracy. In summary, our contributions are threefold:    \begin{itemize}

\item We investigate an unexplored problem: how to obtain a multi-organ segmentation network by combining off-the-shelf single-organ segmentation models. To address this challenge, we propose a baseline approach that can serve as a starting point for further research in this field.

\item We propose a novel model adaptation strategy for adapting off-the-shelf models to target domains, without requiring access to source data or altering the structures of source models. Our adaptation strategy mainly consists of a Label Refinement Module and a Feature Generalization Module, which enhances the model generalization from the perspective of pseudo-labeling and feature space optimization.

\item We propose a novel model ensemble method for aggregating multiple single-organ segmentation models into a tailored multi-organ segmentation model. For this goal, we devise a certainty-aware ensemble function to dynamically compose the teacher model from multiple single-organ segmentation models, and then transfer knowledge from the teacher model to the target multi-organ segmentation model.

\end{itemize}

\section{RELATED WORK}
\label{related_work}

In this section, we briefly review the most related works in
the literature: \textbf{(1)} medical image segmentation with model adaptation (Sec.~\ref{sfda_related}) and \textbf{(2)} distillation and ensemble (Sec.~\ref{de_related}).

\subsection{Medical image segmentation with model adaptation}
\label{sfda_related}
In recent years, several studies proposed the concept of source-free domain adaptation (SFDA)~\cite{kundu2020universal, liang2020we}, which adapts models without access to source data. These methods have been widely used in medical scenarios due to the constraints of data availability. 
For example, Bateson et al.~\cite{bateson2020source} proposed AdaMI, which extracted prior knowledge by adding auxiliary branches in the source models to refine the task. Stan et al.~\cite{stan2021privacy} encoded the source samples into a prototypical distribution, and aligned the distribution to the target domain.
However, these SFDA methods need to alter the structure of the source model with auxiliary branches or additional training tasks, which could not adapt off-the-shelf models that are already well-trained on the source domains.

To address this limitation, some researchers focused on improving segmentation performance only using an off-the-shelf segmentation model, which is known as model adaptation. Existing model adaptation methods reduce domain shifts through either pseudo-labeling or distribution alignment.
In terms of pseudo-labeling methods, Cheng et al.~\cite{chen2021source} proposed the denoised pseudo-labeling (DPL), which filtered out noisy pseudo labels and re-trained the model based on the preserved pseudo labels. 
In terms of distribution alignment methods, Liu et al.~\cite{liu2021source} proposed an adaptive batch-wise normalization statistics adaptation framework to adapt off-the-shelf segmentation models. Bruggemann et al.~\cite{bruggemann2023contrastive} leverages unlabeled pairs of adverse- and normal- condition images to learn condition-invariant features via contrastive learning.

In this paper, we argue that pseudo-labeling and distribution alignment methods are valuable complements of each other. Our experiments have shown that distribution alignment is effective at adapting the feature distribution, while pseudo-label refinement is better at refining the decision boundary. Therefore, we propose to integrate both pseudo-labeling and distribution alignment to improve model adaptation.

\subsection{Distillation and ensemble}
\label{de_related}
Distillation and ensemble is extended from knowledge distillation, which transfers knowledge from teachers to a compact student model. Hinton et al.\cite{hinton2015distilling} were the first to investigate transferring knowledge from a teacher model ensemble to the student. They simply combined soft predictions of teacher models using the averaging operation and adopted KL-divergence loss for knowledge transfer. Malinin et al.\cite{malinin2019ensemble} argued that using the averaging operation for combination harmed the diversity of the models in an ensemble. To address this issue, they used a prior network to estimate output uncertainties of different teacher models, and fused predictions with the guidance of uncertainty. However, the above approaches still depend on labeled datasets, which cannot meet many real-world scenarios where manual labels are unavailable. To solve this problem, Chao et al.\cite{chao2021rethinking} introduced an unsupervised domain adaptation-based ensemble-distillation framework, which was robust to the inconsistency in the scale of the output certainty values and the performance variations among the members in an ensemble. However, this method needed to use adapted models instead of directly using off-the-shelf models, and still had the strict requirement for the label consistency, which is not available for real scenarios.

\section{Method}
\label{method}

\begin{figure*}[t]
\centering
\includegraphics[width=1\textwidth]{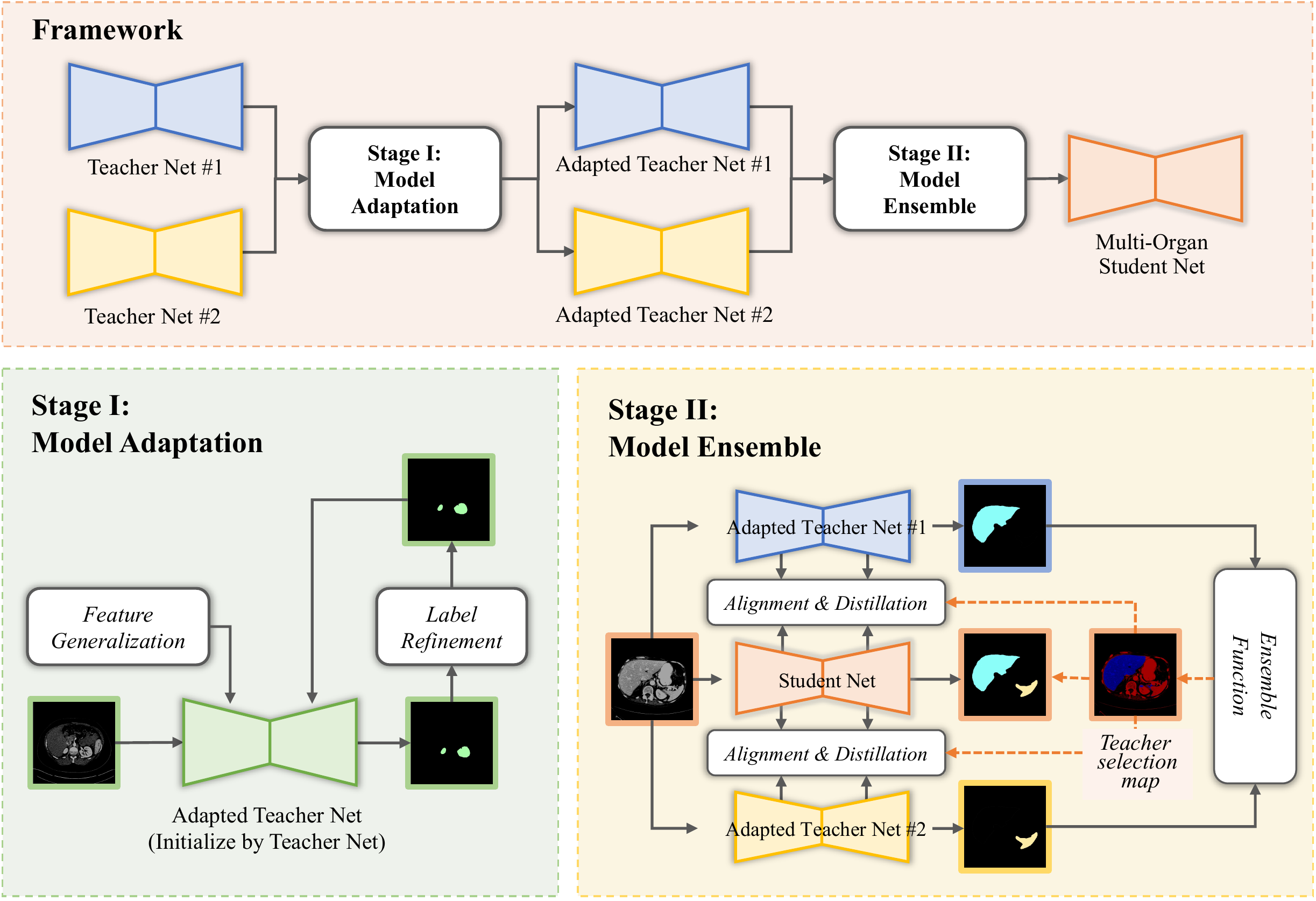}
\caption{A schematic view of our Multi-Model Adaptation framework, which mainly experiences the Model Adaptation and Model Ensemble stage.}
\label{Fig_Overview}
\end{figure*}

In this section, we introduce the proposed Multi-Model Adaptation method from three aspects: First, we give an overview of the framework in Section.~\ref{sec_overview}, including the problem definition and overall pipeline; Then, we delve into the details of the model adaptation strategy in Section.~\ref{sec_ma}; Finally, we provide details of the model ensemble strategy in Section.~\ref{sec_ma}. 

\subsection{Overview}
\label{sec_overview}
\subsubsection{Problem Definition}
Assuming there exist a pool of off-the-shelf single-organ segmentation models, denoted as $\{T_1, T_2, ..., T_n\}$. These models were trained for distinct single-organ segmentation tasks, denoted as $\{\mathcal{K}_1, \mathcal{K}_2, ..., \mathcal{K}_n\}$, and were optimized using different datasets with varying distributions. Let $\mathcal{K}=\bigcup_i K_i$ denote the union of segmentation tasks covered by all models. Our goal is to combine information from all the off-the-shelf models to build a target multi-organ segmentation model for tasks $\mathcal{K}$ that can perform well on an unseen target dataset with a new distribution.

\subsubsection{Overall Pipeline}
The overview of our framework is illustrated in Fig.~\ref{Fig_Overview}. To clearly introduce our method, we consider the simplest scenario: obtaining a dual-organ segmentation model (also referred to as the student net) from two single-organ segmentation models (also referred to as teacher nets). However, our method is not limited to just two teacher models; it can be extended to aggregate any number of single-organ segmentation models.

As shown in Fig.~\ref{Fig_Overview}, the overall training process experiences two stages: the Model Adaptation stage and the Model Ensemble stage. In the Model Adaptation stage, we adapt teacher nets to fit the target data distribution. These adapted models, which generalize well on the target dataset, are called adapted teacher nets. In the Model Ensemble stage, we obtain a tailored student net from the union of adapted teacher nets that can complete two segmentation tasks (the union of tasks of all teacher nets). Below, we delve into the specifics of each stage.

\subsection{Stage-I: Model Adaptation}
\label{sec_ma}

\begin{figure}[t]
\centering  
\subfigure[]{
\begin{minipage}[t]{0.45\linewidth}
\centering
\includegraphics[width=4cm]{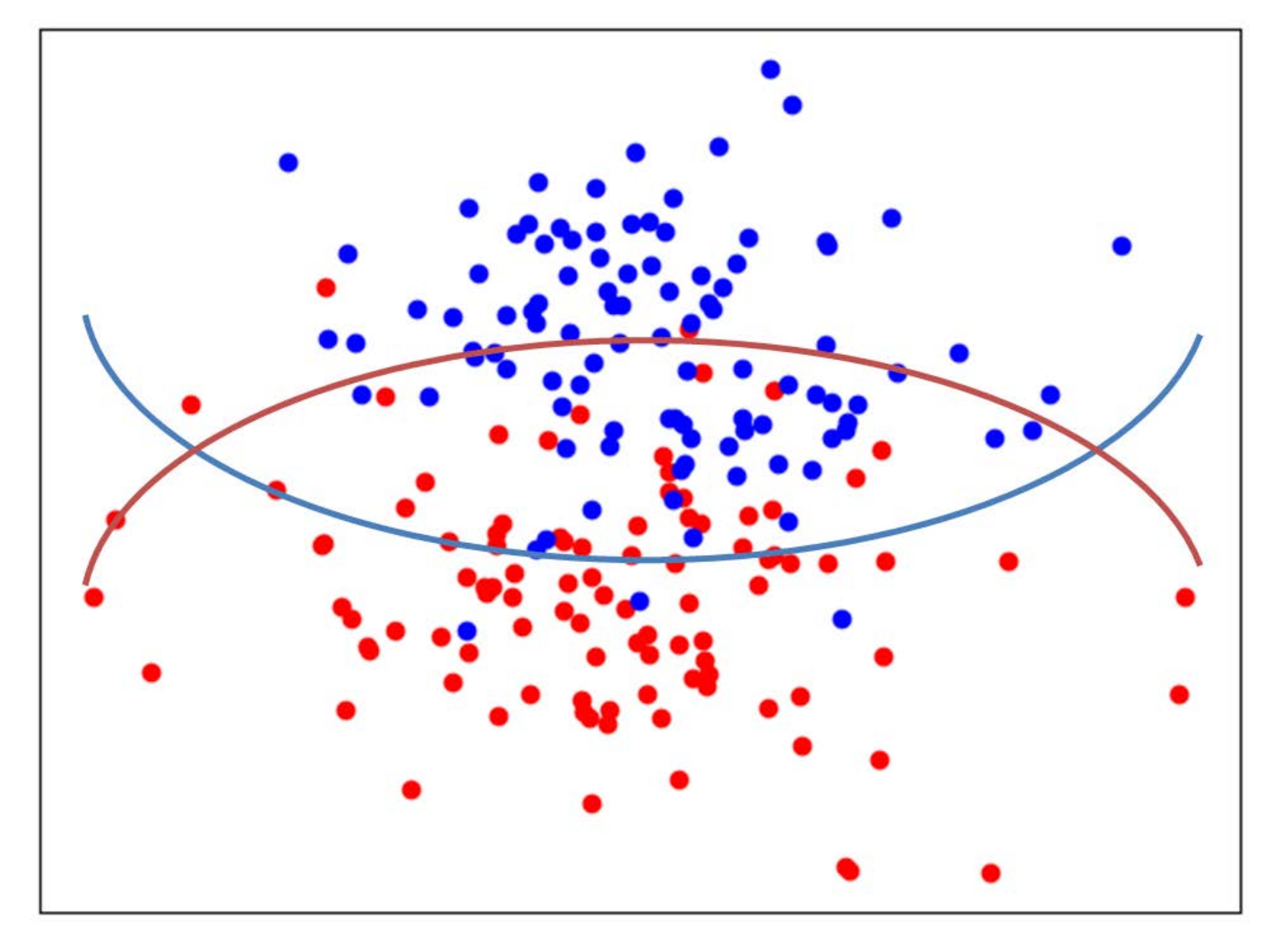}
\end{minipage}%
}
\subfigure[]{
\begin{minipage}[t]{0.45\linewidth}
\centering
\includegraphics[width=4cm]{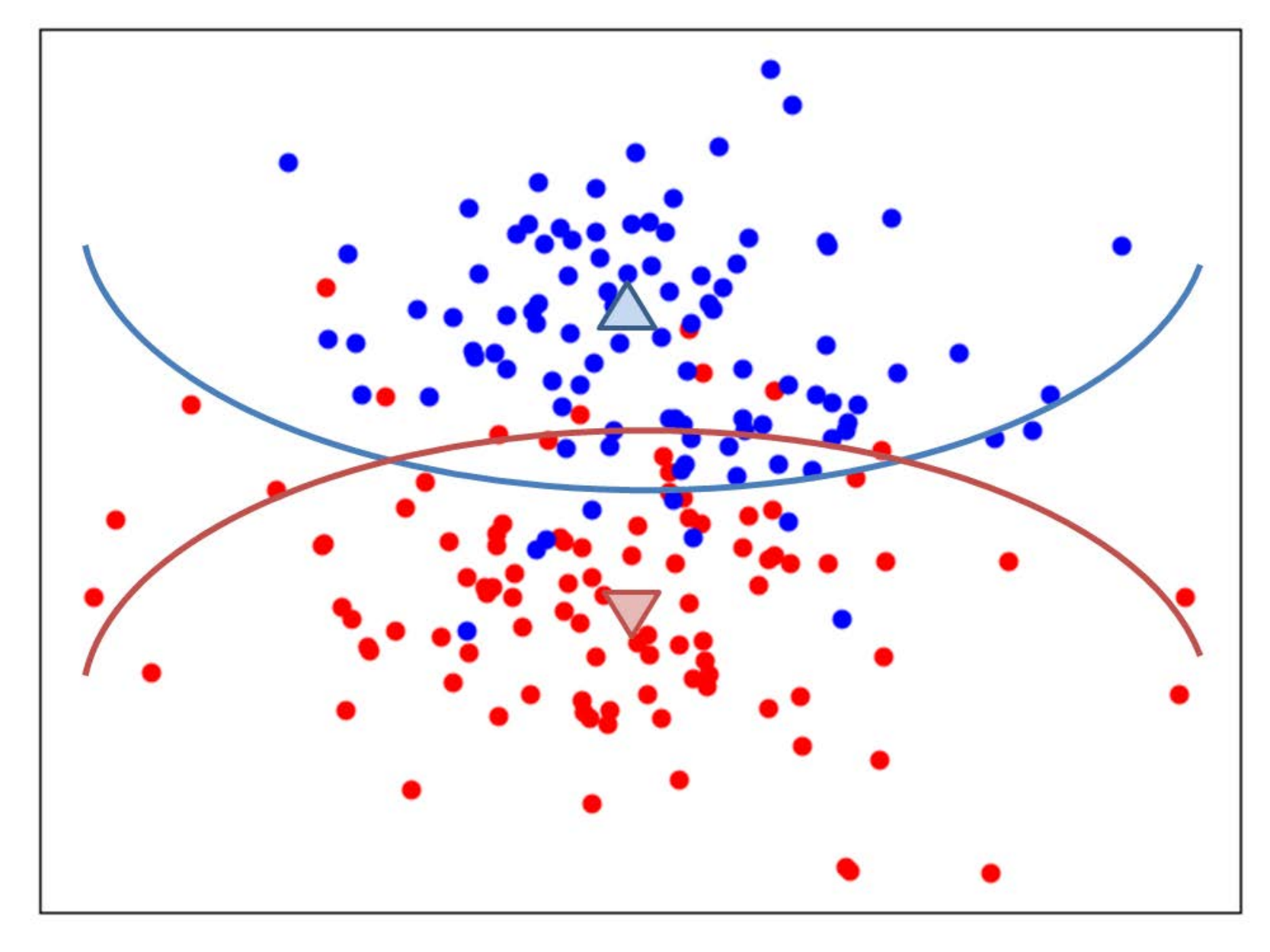}
\end{minipage}%
}
\subfigure[]{
\begin{minipage}[t]{0.45\linewidth}
\centering
\includegraphics[width=4cm]{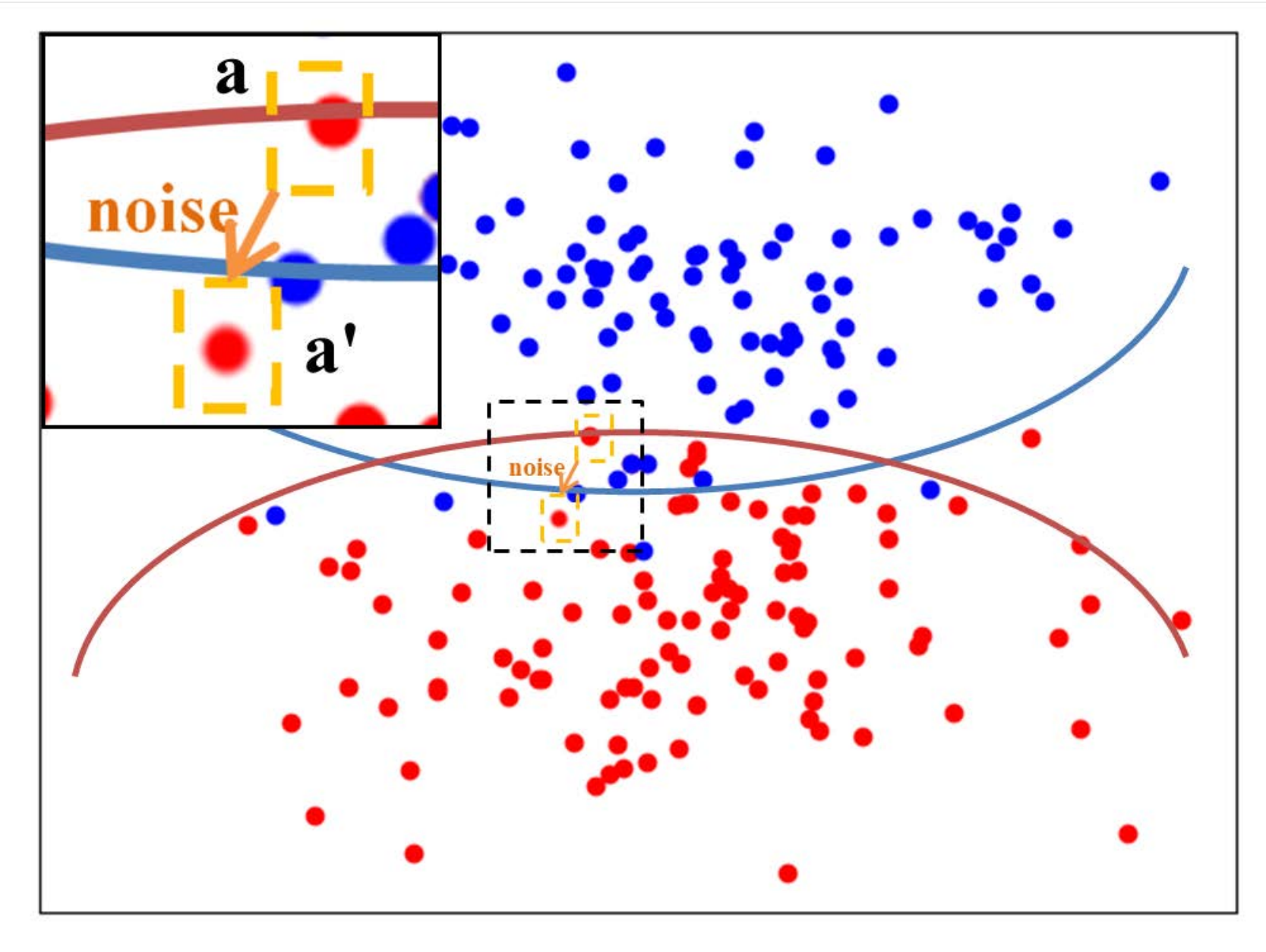}
\end{minipage}%
\label{Fig_Explain_noise}
}
\subfigure[]{
\begin{minipage}[t]{0.45\linewidth}
\centering
\includegraphics[width=4cm]{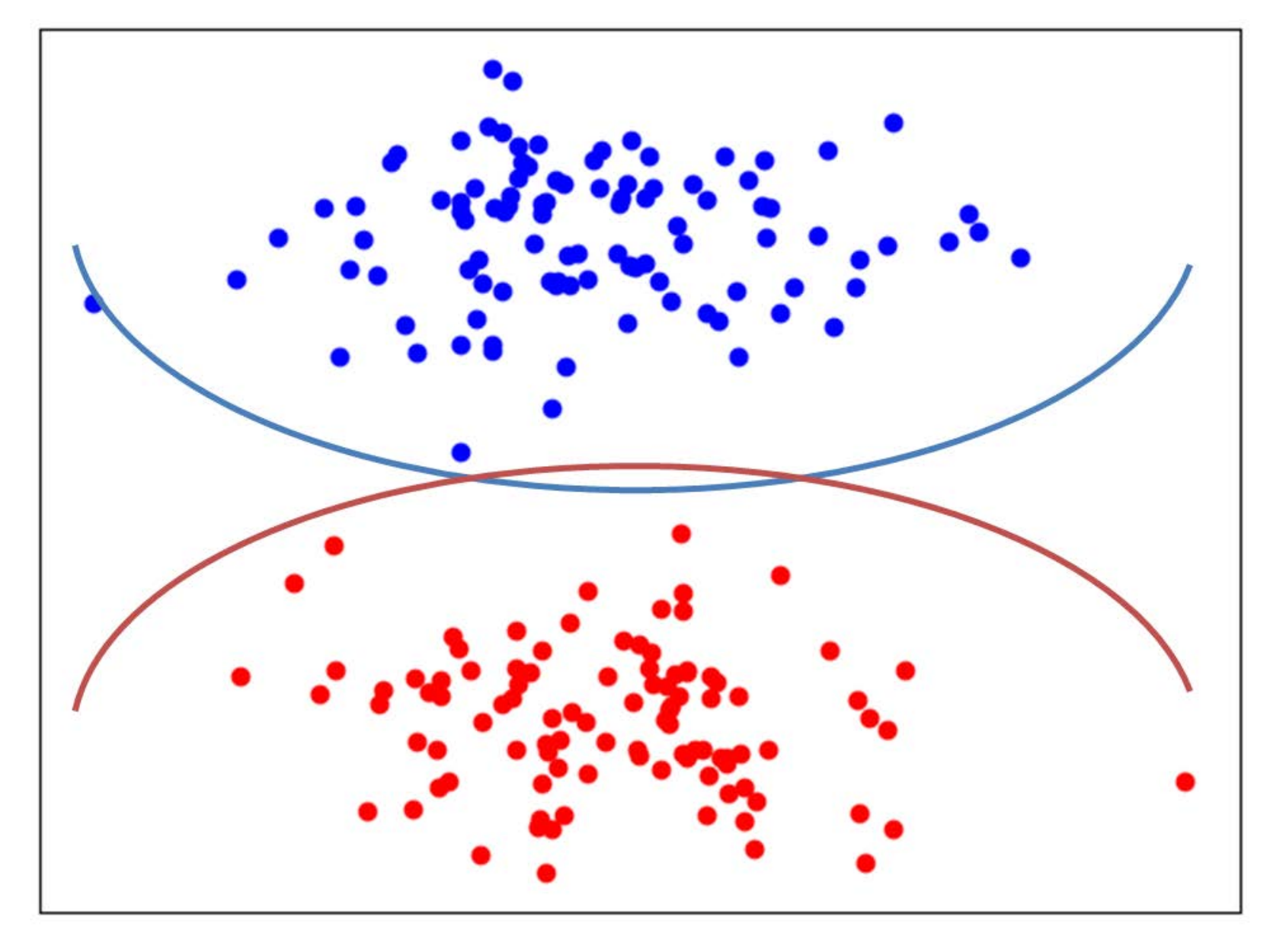}
\end{minipage}%
}
\caption{The schematic of different stages of model adaptation. Panel (a) shows the feature space and decision boundary of the target data before model adaptation. Panel (b) and (c) show the refined decision boundary and feature distribution through LRM and FGM, respectively. By incorporating LRM and FGM, we can obtain the ideal feature space and boundary in panel (d).}
\label{Fig_Feature_Transfer}
\end{figure}

The Model Adaptation stage adapts each teacher net to make it fit the target data distribution. Fig.~\ref{Fig_Feature_Transfer} (a) illustrates the feature distribution and decision boundary on the target domain generated by the teacher net before adaptation. Due to domain shifts between the source dataset and target dataset, the dense classification fails in many pixel locations. Our ultimate objective is to optimize the feature distribution and obtain an accurate decision boundary (see Fig.~\ref{Fig_Feature_Transfer} (d)), such that the teacher net is adapted well to the target domain. To achieve this, we introduce the label refinement module (LRM) and feature generalization module (FGM). Specifically, LRM produces reliable pseudo-labels for the target dataset, which avoids having decision boundaries in high-density areas (see Fig.~\ref{Fig_Feature_Transfer}(b)).
Meanwhile, FGM enforces robust and compact feature distribution by leveraging a consistency loss and information maximization loss (see Fig.~\ref{Fig_Feature_Transfer}(c)). By combining both two modules, we finally obtain a well-adapted teacher net (see Fig.~\ref{Fig_Feature_Transfer} (d)).
Below, we elaborate on the details of LRM and FGM.


\subsubsection{Label Refinement Module}
\label{Sec_label_denoise}
Common pseudo-labeling methods directly used network predictions as pseudo-labels, which inevitably generate error-prone labels for the target domain. As training progresses, these errors accumulate and could significantly degrade model performance. 
Considering this, LRM is proposed to preserve certain confident pseudo-labels and refine low-confident ones based on class prototypes.


Generally, a class prototype is the most representative feature vector that characterizes its category.
For each class $c$ in the target domain, its class prototype $proto_c$ is calculated by combining the feature vector of each pixel:
\begin{equation}
proto_c = \frac{\sum_{x \in X} \delta_c(x) f(x)}{\sum_{x \in X} \delta_c(x)} 
\end{equation}
where $x$ is a pixel location, $\delta_c(x)$ is the probability score for class $c$ (yielded by the Softmax layer of the teacher net), and $f(x)$ decision feature of $x$ yielded by the second-to-last convolutional layer.

Given that a reliable pseudo-label should be similar to its class prototype ~\cite{lee2013pseudo, zhang2021prototypical}, we measure the similarity between the features of each pixel with the prototype using cosine similarity:
\begin{equation}
\label{Equ_sim_distance}
Sim_c(x)= \frac{\left\|(f(x) \cdot proto_c\right\|_{2}}{\left\|(f(x)\right\|_{2} \left\|proto_c\right\|_{2}}
\end{equation}
where $Sim_c(x)$ is the similarity between pixel $x$ and prototypes of class $c$, operator $||*||_2$ means L2 norm.

So far, we can refine pseudo-labels according to the following equations:
\begin{equation}
\hat{y}(x)=\begin{cases}
\arg \min _{c} Sim_{c}(x), & \lambda > \lambda_{th}\\
 y^{\prime}(x), & \lambda \leq \lambda_{th}
\end{cases}
\label{eq_3}
\end{equation}
where $\lambda$ is the entropy measuring the confidence of network output, $\lambda_{th}$ is the entropy threshold which is set according to different segmentation tasks (\emph{e.g.}, 0.1 for liver, 0.4 for spleen, and 0.2 for kidney segmentation considering different organ characteristics), $y^{\prime}(x)$ is the prediction of source model. 
For such confidently classified pixels (with $\lambda \leq \lambda_{th}$), we preserve the original network prediction. Meanwhile, for the remnant low-confident pixels (with $\lambda > \lambda_{th}$), we update their pseudo-labels to the class that is most similar to them using Eq.~\ref{eq_3}.
Then, the refined-pseudo labels are used to fine-tune the teacher net using a cross-entropy loss:
\begin{equation}
\label{equ_ma_ce_loss}
\begin{aligned}
&L_{lrm}(x)=L_{ce}(\hat{y}(x), \delta(x))  
\end{aligned}
\end{equation}
where $L_{ce}$ denotes the cross-entropy loss; $\delta(x)$ is the Softmax output; and $\hat{y}(x)$ is the refined pseudo label. 

Intuitively, the LRM modifies the original decision boundaries (Fig.~\ref{Fig_Feature_Transfer} (a)) by updating the classification of each pixel. This avoids the decision boundary in high-density regions on the target domain (Fig.~\ref{Fig_Feature_Transfer} (b)) and reduces the ambiguity of predictions.

\subsubsection{Feature Generalization Module}
\label{Uncentainty}
The FGM further improves the model adaptation from the perspective of feature space optimization. It encourages a more compact feature distribution and reduces the class conditional distribution overlap (see Fig.~\ref{Fig_Feature_Transfer}(c)).
We achieve this by introducing two loss functions, including a consistency loss and an information maximization loss.

Specifically, the consistency loss aims to enforce the invariance of the model’s predictions over small perturbations introduced to the network. To apply perturbations to the hidden representation, 
we introduce multiple auxiliary decoders that use dropout, while the original main decoder does not use dropout~\cite{ouali2020semi}.
The consistency constraint is then imposed between predictions of the main decoder and auxiliary decoders:
\begin{equation}
\label{equ_ma_un_loss}
\begin{aligned}
&\mathcal{L}_{con}(x) = \frac{1}{K} \sum_{k=1}^{K}\left(\hat{\delta}^{k}(x)-\delta(x)\right)^{2}
\end{aligned}
\end{equation}
where $K$ is the number of auxiliary decoders that is set to 4, $\delta(x)$ is the output of the main decoder, and $\hat{\delta}^k(x)$ is the output of $k^{th}$ auxiliary decoder. 
 
The information maximization (IM) loss is designed to reduce uncertainty while ensuring the diversity of categories~\cite{liang2020we}, which consists of an entropy loss item $L_{ent}$~\cite{krause2010discriminative} and a diversity loss item $L_{div}$~\cite{shi2012information}:
\begin{equation}
\label{equ_ma_im_loss}
\begin{aligned}
\mathcal{L}_{i m}(x) &=\mathcal{L}_{e n t}(x)+\beta \mathcal{L}_{d i v}(x)
\end{aligned}
\end{equation}
where the trade-off coefficient $\beta$ is impirically set to 0.1.

Finally, the overall objective of the Model Adaptation stage is formulated as:
\begin{equation}
L_{ma}(x) = L_{lrm}(x) + \lambda_{con}L_{con}(x) + \lambda_{im}L_{im}(x)
\end{equation}
where $\lambda_{con}$ and $\lambda_{im}$ are trade-off coefficients, which are set to 0.1 and 1 respectively.

\subsection{Stage-II: Model Ensemble}
\label{sec_me}

The Model Ensemble stage aims to obtain a tailored student net for multi-organ segmentation from the union of adapted teacher nets, whose architecture is shown in Fig.~\ref{Fig_Overview}. 
In this stage, constructing an ensemble function that indicates the aggregation criteria is the key issue, which can greatly affect the effectiveness of the model ensemble. Then, we can aggregate multiple teacher models into a student model based on the ensemble function from two aspects: label aggregation and feature aggregation.


\subsubsection{Ensemble Function}
To combine effective information from adapted teacher nets, we propose the certainty-aware ensemble function, which dynamically selects the best teacher model for each pixel position by computing a certainty-based teacher selection map. 


\emph{Definition of Model Certainty:} 
Generally, entropy can be used to quantify the certainty of teacher models, in which lower entropy values indicate higher certainty. To ensure comparable ranges of entropy values across different teacher models,
we regularize the entropy values using mean normalization, which is formulated as:
\begin{equation}
\label{Equ_Uniform}
\begin{aligned}
    H(x; t_i) = -\sum_{c} p_{c} (x;t_i)log p_{c} (x;t_i) \\
    H_{norm}(x; t_i) = \frac{H(x; t_i) - \mu_{y(x;t_i), t_i}}{\sigma_{y(x;t_i), t_i}} 
\end{aligned}
\end{equation}
where $t_i$ is the $i^{th}$ teacher model, $c$ is the class, $p_{c} (x;t_i)$ is the probability of pixel $x$ belonging to class $c$ yielded by the $i^{th}$ teacher model, $y(x;t_i)$ is the prediction result of $i^{th}$ teacher model for pixel $x$, $\mu$ and $\sigma$ is the mean and standard deviation. The normalized entropy value $H_{norm}$ is regarded as the certainty of a model.

\emph{Certainty-Based Teacher Selection Map:}
Considering each teacher model for single-organ segmentation cannot provide decision-making information for other segmentation tasks that have not been learned, we exclude this part of the teacher net before selecting the best teacher net. For each pixel $x$, we prepare the pre-selected teacher models and define the set as $A(x)$, which is formulated as:
\begin{equation}
\label{equ_preselected}
A(x)= \begin{cases}
\text (i) A_o, & \text { if } \forall t_i, y(x;t_i)=0 \\
\text (ii) A_o \backslash \bigcup A_{c'}, c' \in D_c, & \text { otherwise }
\end{cases}
\end{equation}
where $A_c$ is the set of teacher nets that have no information about class $c$, $A_o$ is the set of all teacher nets and $D_c = \{c | \forall t_i \in A_c, y(x;t_i)=0 \}$ is the set of categories that can not be identified as $c$ (all teacher models $t_i$ in $A_c$ predict pixel $x$ as background). In Eq.~\ref{equ_preselected}, (i) is the condition when pixel $x$ is unlabeled by all teacher models and (ii) is the condition when there exist conflicts and we need to discard the teacher nets that lack information about the category of pixel $x$. 

Then, the teacher model with the highest certainty among $A(x)$ is chosen to generate the teacher selection map:
\begin{equation}
\label{Equ_Teacher_selection}
t_{se}(x) = \mathop{\arg\min}\limits_{t_i \in A(x)} H_{norm}(x;t_i)
\end{equation}
where $t_{se}$ denotes the selected teacher model for a pixel location $x$. For an image sample, different pixel locations correspond to different selected teachers.

\subsubsection{Label Aggregation}
To aggregate knowledge from multiple adapted teacher nets, we first aggregate single-organ predictions into multi-organ pseudo-labels based on the teacher selection map for self-training. We denote $y_{aggr}$ as the aggregated multi-organ pseudo-label and express it as follows:
\begin{equation}
y_{aggr}(x)= y(x;t_{se}(x))
\end{equation}

Then, the aggregated pseudo-labels are used to train the student net using a cross-entropy loss:
\begin{equation}
\label{equ_loss_cl}
L_{la}=L_{ce}(y_{aggr}(x),p_s(x))
\end{equation}
where $p_{s}(x)$ is the output of student model at the pixel $x$.


\subsubsection{Feature Aggregation}
To help the student net learn the advantages of each network, we also aggregate the knowledge from multiple adapted teachers under the guidance of the teacher selection map. To achieve this, 
we first project the teacher net and student net into the same feature space, which is implemented by a learnable convolution layer with $3\times 3$ kernel. The projection can be formulated as:
\begin{equation}
\label{equ_fx}
F_{s}^{\prime}=Conv_{3\times3}(F_{s}), \ F_{t_i}^{\prime}=Conv_{3\times3}(F_{t_i})
\end{equation}
where $F_s$ ($F_{t_i}$) represent the original features of student model $s$ (teacher model $t_i$) and $F_s^{\prime}$ ($F_{t_i}^{\prime}$) represent the projected features of student model $s$ (teacher model $t_i$).

After that, we distill the knowledge from the selected teacher net to the student net. Specifically, we minimize the distance between the projected feature of the student net and the selected teacher net (based on the teacher selection map, Eq.~\ref{Equ_Teacher_selection}) via the L2 loss. And the feature distillation loss $L_{fa}$ of learned teacher model $i$ is formulated as:

\begin{gather}
    \label{equ_Lkbt}
    \mathcal{L}_{fa}^{(i)}(x) = M_i(x) * \left\|Upsample[F_{s}^{\prime}](x)-Upsample[F_{t_i}^{\prime}](x)\right\|^{2}_2 \nonumber \\
    M_i(x)= \begin{cases}
    \text 1, & \text { if } t_{se}(x)=t_i  \\
    \text 0, & \text { otherwise }
    \end{cases}
\end{gather}
where $Upsample[F_s^{\prime}(x)]$ ($Upsample[F_{t_i}^{\prime}(x)]$) means up-sampling the projected features of student model $s$ (teacher model $t_i$) to the size of teacher selection mask. In our experiments, all the teacher nets and the student net follow the architecture of DeepLabv3+~\cite{chen2017deeplab}. And we conduct feature aggregation at the fourth encoding block (low-level feature) and the last convolutional layer (high-level feature) of DeepLabv3+~\cite{chen2017deeplab}.

Finally, the overall loss function for the Model Ensemble stage is defined as follows:
\begin{equation}
L_{me}(x) = L_{la}(x) + \lambda_{fa}\sum_{i=1}^{n} {L}_{fa}^{(i)} (x)
\end{equation}
where $n$ is the number of teacher models ($n=2$ for the simplest scenario), $\lambda_{fa}$ is the trade-off coefficient that is set to 0.001.

\section{Experiments and Results}
\label{experiments}

\subsection{Experimental Setup}
\subsubsection{Materials}
In this study, to validate the effectiveness of our method, we conducted extensive experiments on four abdominal segmentation datasets, including the LiTS dataset, RS dataset, BTCV dataset, and SRRSHA dataset. 

The \textbf{LiTS dataset}~\cite{bilic2019liver} is a public dataset consisting of 131 CT scans that have been annotated to identify both liver and liver tumors. To create a dataset that exclusively focuses on the liver, we merged the liver and tumor labels by adjusting the label values of the tumor areas to match those of the liver areas.

The \textbf{RS dataset}~\cite{dong2015segmentation} is an in-house spleen segmentation dataset collected by Ritsumeikan University, which includes 51 CT scans with spleen delineations.

The \textbf{BTCV dataset}~\cite{xu2016evaluation} is a public dataset including 30 CT scans with 13 organs annotated. In our experiments, we only use the liver and spleen labels. 

The \textbf{SRRSHA dataset} is an in-house dataset collected by Sir Run Run Shaw Hospital, which includes 277 MRI scans with 4 abdomen organs annotated. In this study, only liver and spleen labels are used. 

\begin{table}[t]
\renewcommand\tabcolsep{2.8pt}
\renewcommand\arraystretch{1.1}
\centering
\scriptsize
\caption{A summary description of the used datasets.}
\label{tab_dataset}

\begin{tabular}{ccccccc}
\toprule[1pt]
\textbf{Dataset} & \textbf{Modality} & \textbf{Scans} & \textbf{Organ} & \textbf{Axis} & \textbf{Size} & \textbf{Spacing} \\
\hline
\multirow{3}{*}{LiTS} & \multirow{3}{*}{CT} & \multirow{3}{*}{131} & \multirow{3}{*}{Liver} & X & 512 & 0.63–1.00 \\
& & & & Y & 512 & 0.63–1.00 \\
& & & & Z & 75–987 & 0.70–5.00 \\
\hline
\multirow{3}{*}{RS} & \multirow{3}{*}{CT} & \multirow{3}{*}{51} & \multirow{3}{*}{Spleen} & X & 512 & 0.63–1.00 \\
& & & & Y & 512 & 0.63–1.00 \\
& & & & Z & 75–987 & 0.70–5.00 \\
\hline
\multirow{3}{*}{BTCV} & \multirow{3}{*}{CT} & \multirow{3}{*}{30} & \multirow{3}{*}{\tabincell{c}{Liver, spleen and \\ other 11 organs}} & X & 512 & 0.63–1.00 \\
& & & & Y & 512 & 0.63–1.00 \\
& & & & Z & 75–987 & 0.70–5.00 \\
\hline
\multirow{3}{*}{SRRSHA} & \multirow{3}{*}{MRI} & \multirow{3}{*}{277} & \multirow{3}{*}{\tabincell{c}{Liver, spleen and\\ other 2 organs}} & X & 512 & 0.63–1.00 \\
& & & & Y & 512 & 0.63–1.00 \\
& & & & Z & 75–987 & 0.70–5.00 \\
\toprule[1pt]
\end{tabular}

\end{table}

Table ~\ref{tab_dataset} shows the details of the four datasets.
LiTS and RS datasets are reserved for source datasets. We randomly split each of them into training/validation/testing sets with a fixed ratio of 60\%:10\%:30\% and use them to train the single-organ segmentation teacher nets. Meanwhile, BTCV and SRRSHA datasets are reserved for target datasets. We randomly split each of them into training/testing sets with a fixed ratio of 70\%:30\% and use them to test the student net. \textsl{(Note that there is no validation set for target datasets since our task is unsupervised model adaptation.)} For preprocessing, all CT slices are truncated into the range of [-250, 250] Hu to eliminate irrelevant tissues; Meanwhile, for MR images, we clipped the 35\% highest intensity values to eliminate irrelevant details. 

\subsubsection{Competing Methods}
Since we are the first to solve the Multi-Model Adaptation problem, there exists no similar work that learns a multi-organ segmentation model from the union of off-the-shelf single-organ segmentation models. Therefore, we compare our method with relevant domain adaptation methods:

\textbf{\textsl{(1) AdvEnt}}~\cite{vu2019advent}, a popular unsupervised domain adaptation benchmark approach that encourages entropy consistency between the source domain and the target domain;

\textbf{\textsl{(2) AdaMI}}~\cite{bateson2020source}, a source-free domain adaptation approach that generates class-ratio prior via an auxiliary network in the source domain and refines the segmentation mask guided by the class-ratio prior in the target domain;

\textbf{\textsl{(3) DPL}}~\cite{chen2021source}, a model adaptation approach that generates more discriminative and less noisy supervision by pixel-level denoising with uncertainty estimation and class-level denoising with prototype estimation;

\textbf{\textsl{(4) MDAN}}~\cite{zhao2018adversarial}, a multi-source domain adaptation approach that distinguishes the pair of the source domain and target domain by adding k domain classifiers.

Among these competing methods, only MDAN can obtain a multi-organ segmentation model from single-orange segmentation models, while other methods merely obtain adapted single-organ segmentation methods. But, it should be noted that MDAN needs access to source domain data.
Besides, AdvEnt, AdaMI, and MDAN are strong competing methods since they utilize more source domain information by either accessing source domain data or altering source structures of source models. However, our method is completely based on off-the-shelf model adaptation, which neither uses source domain data nor alters source model structures.

\subsubsection{Evaluation Metrics} 
To quantitatively evaluate the methods, we adopt the Dice similarity coefficient (DSC) and average surface distance (ASD) as the metrics. Specifically, DSC is for pixel-wise accuracy measurement, and ASD is for boundary segmentation evaluation. The higher DSC and lower ASD indicate better performance.

\subsubsection{Implmetation Details}
In this study, all methods are implemented using Pyotorch 1.5.0 and deployed on an NVIDIA GTX 2080 GPU.
To ensure the fairness of the comparison, all methods adopt DeepLabv3+~\cite{chen2017deeplab} with MobileNetV2~\cite{sandler2018mobilenetv2} as the network backbone. We adopt the Adam optimizer for training with the momentum of 0.9 and 0.99, and a learning rate of 2e-3. The training epoch number of the model adaption stage is set to 100 with a batch size of 4; Meanwhile, the training epoch number of the Model Ensemble stage is set to 200 with a batch size of 2.

\begin{figure*}[t]
\centering
\includegraphics[width=\textwidth]{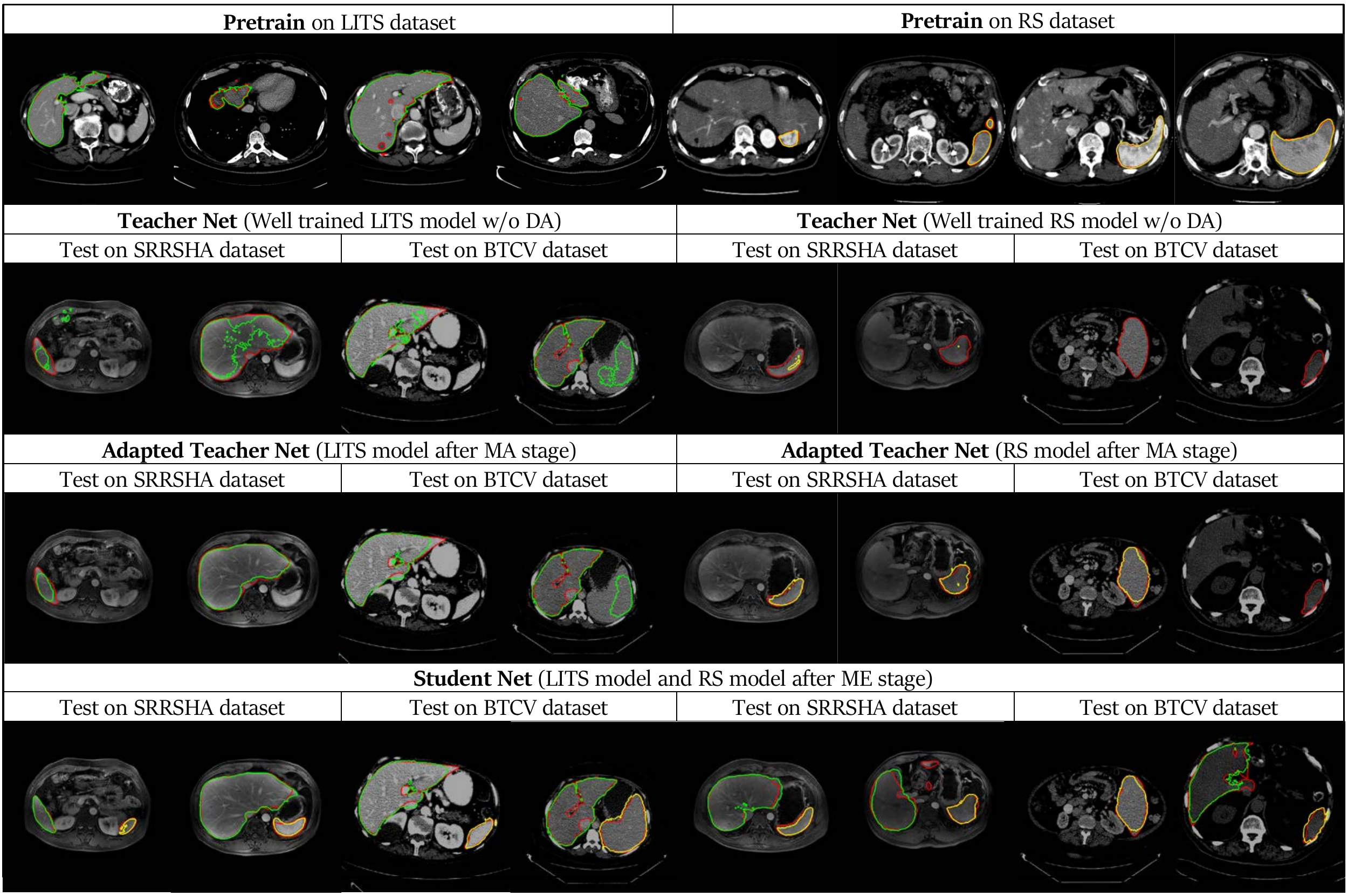}
\caption{Visual results produces by each stage of our method on four abdomen datasets. The ground truths are delineated in \textcolor[RGB]{178, 34, 34}{\textbf{red}}, the prediction results of liver segmentation are delineated in \textcolor[RGB]{34, 139, 34}{\textbf{green}} and the prediction results of spleen segmentation are delineated in \textcolor[RGB]{255, 193, 27}{\textbf{yellow}}.}
\label{Fig_Phase_Results_A}
\end{figure*}

\subsection{Results of the Proposed Method}

\begin{table}[t]
\scriptsize
\renewcommand\tabcolsep{7.5pt}
\renewcommand\arraystretch{1.1}
\centering
\caption{Quantitative analysis and comparison of well-trained models (w/o domain adaptation) verified on different domains.}
\label{tab_domain_gap}

\begin{tabular}{cccccc}
\toprule[1pt]
\multirow{2}{*}{\tabincell{c}{\textbf{Source} \\ \textbf{Domain}}} & \multirow{2}{*}{\tabincell{c}{\textbf{Target} \\ \textbf{Domain}}} & \multicolumn{2}{c}{\textbf{Liver}} & \multicolumn{2}{c}{\textbf{Spleen}} \\
& & DSC & ASD & DSC & ASD\\
\hline

\multirow{3}{*}{LiTS} & LiTS & 0.9495 & 2.93 & - & - \\
& SRRSHA & 0.7551 & 4.89 & - & - \\
& BTCV & 0.8969 & 4.20 & - & - \\
\hline

\multirow{3}{*}{RS} & RS & - & - & 0.9670 & 0.18 \\
& SRRSHA & - & - & 0.3484 & 5.86 \\
& BTCV & - & - & 0.6973 & 10.11 \\
\toprule[1pt]
\end{tabular}

\end{table}

\begin{figure}[t]
\centering
\includegraphics[width=\textwidth]{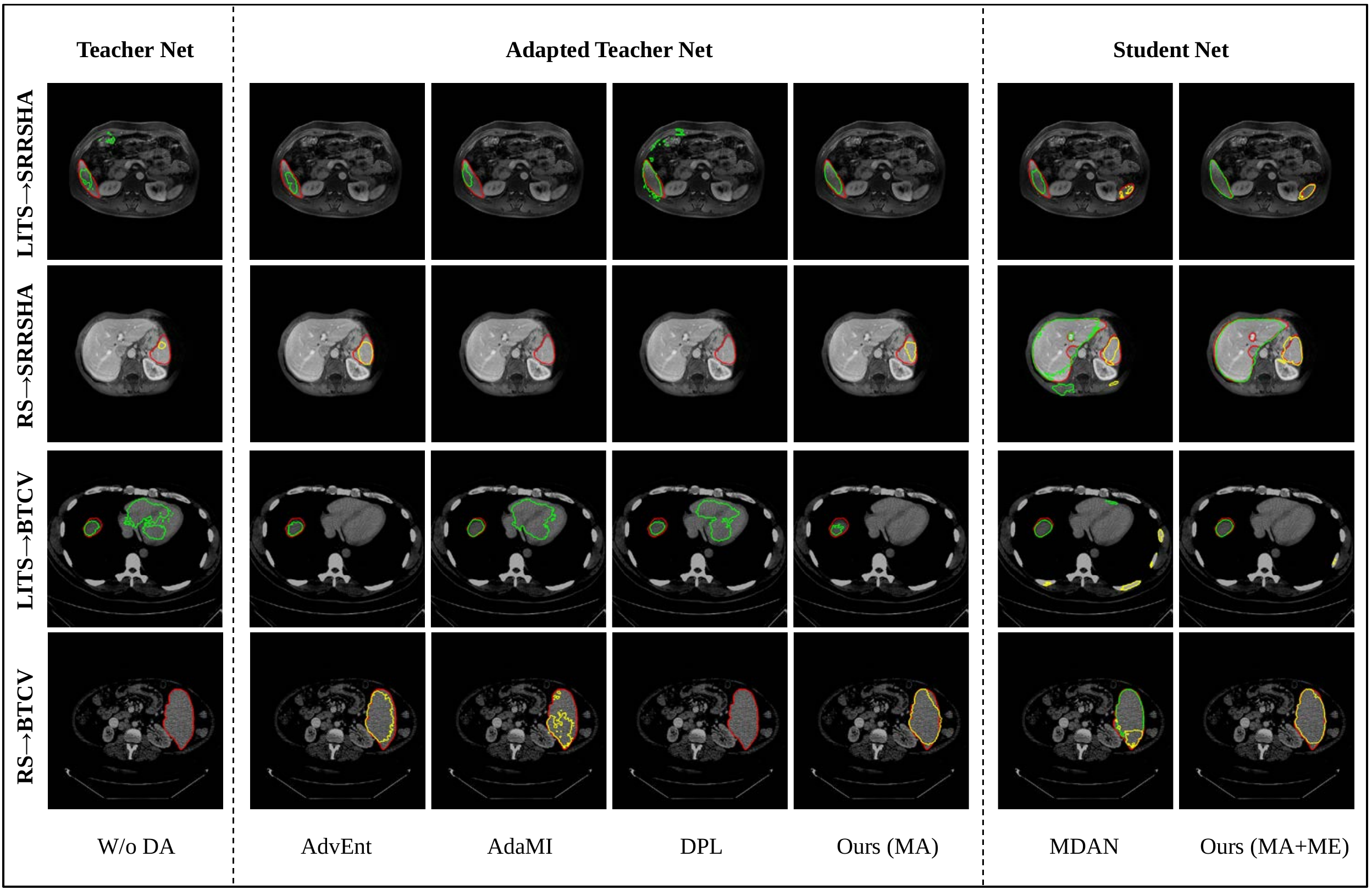}
\caption{Visual comparison results between our proposed MA and different domain adaptation methods verified on Abdomen Dataset. The ground truths are delineated in \textcolor[RGB]{178, 34, 34}{\textbf{red}}, the prediction results of liver segmentation are delineated in \textcolor[RGB]{34, 139, 34}{\textbf{green}} and the prediction results of spleen segmentation are delineated in \textcolor[RGB]{255, 193, 27}{\textbf{yellow}}.} 
\label{Fig_Abdomen_SOTA}
\end{figure}

\begin{table}[t]
\scriptsize
\renewcommand\tabcolsep{2.6pt}
\centering
\caption{Quantitative analysis and comparison of different methods verified on different Datasets. ``SOS" represents single-organ segmentation, ``MOS" represents multi-organ segmentation. ``\checkmark" means the method accesses the source dataset; ``\st{\checkmark}" means the method alters the structures of source teacher models; ``$\times$" means the method neither needs the source data nor alters source model structures.}
\label{tab_result_1}

\begin{tabular}{cccc|cccc|cccc}
\toprule[1pt]

&{\multirow{3}{*}{\textbf{Method}}} & \multirow{3}{*}{\tabincell{c}{\textbf{Source} \\ \textbf{Access}}} & \multirow{3}{*}{\tabincell{c}{\textbf{Source} \\ \textbf{Domain}}} &
\multicolumn{4}{c|}{{\textbf{SRRSHA Dataset}}} &
\multicolumn{4}{c}{{\textbf{BTCV Dataset}}} \\
\cline{5-12}

& & & & \multicolumn{2}{c}{\textbf{Liver}} & \multicolumn{2}{c|}{\textbf{Spleen}} &
\multicolumn{2}{c}{\textbf{Liver}} & \multicolumn{2}{c}{\textbf{Spleen}}
\\
& & & & DSC & ASD & DSC & ASD & DSC & ASD & DSC & ASD\\
\hline

\tabincell{c}{\textbf{Upper-} \\ \textbf{Bound}}&{Supervised} & $\times$ & \tabincell{c}{SRRSHA \\ / BTCV}  & 94.41 & 1.38 & 87.67 & 0.56 & 95.93 & 1.46 & 95.38 & 0.28 \\
\hline
\multirow{2}{*}{\tabincell{c}{\textbf{Lower-} \\ \textbf{Bound}}} & \multirow{2}{*}{{Teacher nets}} & \multirow{2}{*}{$\times$}& LiTS & 75.51 & 4.89 & - & -& 89.69 & 4.20 & - & - \\
& & & RS & - & - & 34.84 & 5.86 & - & - & 69.73 & 10.11 \\
\hline

\multirow{8}{*}{\textbf{SOS}} & \multirow{2}{*}{AdvEnt~\cite{vu2019advent}} & \multirow{2}{*}{\checkmark} & LiTS & 84.38 & 3.70 & - & - & 91.54 & 2.62 & - & - \\
& & & RS & - & - & 76.77 & 1.91 & - & - & 85.49 & 2.72 \\
& \multirow{2}{*}{AdaMI~\cite{bateson2020source}} & \multirow{2}{*}{\st{\checkmark}} & LiTS & 62.92 & 5.68 & - & - & 89.64 & 4.50 & - & - \\
& & & RS & - & - & 55.04 & 4.66 & - & - & 83.60 & 3.84 \\
\cline{2-12}

& \multirow{2}{*}{DPL~\cite{chen2021source}} & \multirow{2}{*}{$\times$}\multirow{2}{*}{} & LiTS & 80.85 & 4.21 & - & - & 89.97 & 4.91 & - & - \\
& & & RS & - & - & 73.79 & 2.47 & - & - & 71.65 & 5.35 \\
& \multirow{2}{*}{MA(ours)} &\multirow{2}{*}{$\times$}\multirow{2}{*}{}& LiTS & 85.16 & 3.21 & - & - & 90.64 & 3.65 & - & - \\
& & & RS & - & - & 79.91 & \textbf{1.51} & - & - & 85.79 & 2.14 \\
\hline

\multirow{2}{*}{\textbf{MOS}} & MDAN~\cite{zhao2018adversarial} & \checkmark & both & 88.24 & 3.73 & \textbf{81.34} & 2.61 & \textbf{93.27} & 3.43 & 80.03 & 5.08 \\
& Ours & {$\times$} & both & \textbf{88.59} & \textbf{2.03} & 80.78 & 1.74 & 91.67 & \textbf{2.14} & \textbf{89.69} & \textbf{1.46} \\
\toprule[1pt]
\end{tabular}

\end{table}

In this section, we present qualitative and quantitative results produced by our proposed method. Fig.~\ref{Fig_Phase_Results_A} shows examples of results produced by different stages of our method on different datasets. 

As observed, although performed well on the source LiTS and RS datasets (as shown in the first row in Fig.~\ref{Fig_Phase_Results_A}), these well-trained models show poor performance on the target BTCV and SRRSHA datasets (see the second row in Fig.~\ref{Fig_Phase_Results_A}). This phenomenon is caused by the domain shifts between the data acquired under different conditions. Evaluated by quantitative analysis, the well-trained liver segmentation model achieves 94.95\% in DSC when tested on the source LiTS dataset, and the well-trained spleen segmentation model achieves 96.70\% in DSC when tested on the source RS dataset.
However, the liver model only achieves 75.51\% in DSC (-19.44\%) on the SRRSHA dataset and 89.69\% (-5.26\%) on the BTCV dataset, and the spleen model only achieves 34.84\% (-61.86\%) in DSC on the SRRSHA dataset and 69.73\% (-26.97\%) on the BTCV dataset (see Table~\ref{tab_domain_gap}). 
Note that a more severe performance drop is observed in cross-modal (CT to MRI) segmentation, i.e., LiTS to SRRSHA and RS to SRRSHA, due to larger domain gaps.

Therefore, the first step in imposing off-the-shelf models is to adapt each well-trained model to the target datasets. To solve this problem, our model adaptation (MA) stage fine-tunes the models without access to corresponding source data. Comparing the second row (before adaptation) and third row (after adaptation) in Fig.~\ref{Fig_Phase_Results_A}, obvious false positives in the first column and false negatives in the sixth column are corrected after model adaptation. More specifically, on the SRRSHA dataset, the dice scores are improved from 75.51\% to 85.16\% (outperforms unadapted teacher net by +9.65\%) for liver segmentation, and from 34.84\% to 79.91\% (+45.07\%) for spleen segmentation. On the BTCV dataset, the dice scores are improved from 89.69\% to 90.64\% (+0.95\%) for liver segmentation, and from 69.73\% to 85.79\% (+16.06\%) for spleen segmentation. 


Finally, to learn a single multi-organ segmentation model from the union of adapted single-organ segmentation models, we introduce the Model Ensemble stage. In this stage, the student net can segment the liver and spleen simultaneously without any harm to performance on both liver and spleen segmentation tasks. Quantitatively, the whole framework achieves the DSC of 88.59\% (outperforms the adapted teacher net by +3.43\%) and 80.78\% (+0.87\%) for liver and spleen segmentation tasks on the SRRSHA dataset, and achieves 91.67\% (+1.03\%) and 89.69\% (+3.90\%) for liver and spleen segmentation on the BTCV dataset.
The improved results demonstrate that our method can effectively acquire essential knowledge from two single-organ segmentation models and resolve the conflicts between the two models. The visual results shown in the fourth row in Fig.~\ref{Fig_Phase_Results_A} are consistent with the quantitative results.

\subsection{Comparison with Domain Adaptation Methods}


In this section, we conduct comprehensive comparison experiments between our method and competing methods.
Table~\ref{tab_result_1} shows the quantitative results of our proposed framework and other methods, in which we also include the lower-bound method (teacher nets before domain adaptation) and the upper-bound method (fully supervised segmentation). The ``Source Access'' column indicates what type of source domain information is used by the methods. Specifically, ``\checkmark" means the method accesses the source dataset; ``\st{\checkmark}" means the method alters the structures of source teacher models; ``$\times$" means the method only uses the off-the-shelf source models, which neither needs the source data nor alters source model structures. It is observed that all the competing methods can improve the performance over the lower-bound method and our proposed method achieves the best performance in most segmentation tasks. Note that our method does not need to access source data, performing even better than AdvEnt and MDAN that use information from source domains (Fig.~\ref{Fig_Abdomen_SOTA}, Table~\ref{tab_result_1}). The underlying reason is that Advent and MDAN mainly focus on feature-level alignment, which only aligns high-level information but ignores pixel-level information.

\subsection{Ablation study}
\label{ablation_study}

In this section, we specifically discuss the rationality of each module in our proposed method on the BTCV dataset. 

\subsubsection{Contribution of Each Component}
\label{component}

\begin{table}[t]
\scriptsize
\renewcommand\tabcolsep{2.5pt}
\renewcommand\arraystretch{1.1}
\centering
\caption{Quantitative Analysis of each component verified on the BTCV Dataset. The whole method experiences two stages: the model adaptation (MA) stage and model ensemble (ME) stage. ``LRM" represent the label refinement module, ``FGM" represents the feature generalization module, ``LA" represents the label aggregation, and ``FA" represents the feature aggregation.}
\label{tab_component}

\begin{tabular}{ccccccc}
\toprule[1pt]

\multicolumn{2}{c}{\multirow{2}{*}{\textbf{Method}}}  & \multirow{2}{*}{\tabincell{c}{\textbf{Source} \\ \textbf{Domain}}} & \multicolumn{2}{c}{\textbf{Liver}} & \multicolumn{2}{c}{\textbf{Spleen}} \\
& & & DSC & ASD & DSC & ASD\\
\hline

\multirow{2}{*}{\tabincell{c}{\textbf{Teacher} \\ \textbf{Net}}} & \multirow{2}{*}{\tabincell{c}{Baseline\\(w/o domain adaptation)}} & LiTS & 0.8969 & 4.20 & - & - \\
& & RS & - & - & 0.6973 & 10.11 \\
\hline

\multirow{4}{*}{\tabincell{c}{\textbf{Adapted} \\ \textbf{Teacher} \\ \textbf{Net}}} & \multirow{2}{*}{LRM} & LiTS & 0.9016 & 3.97 & - & - \\
& & RS & - & - & 0.8391 & 4.20 \\
& \multirow{2}{*}{\tabincell{c}{MA \\ (LRM+FGM)}} & LiTS & 0.9064 & 3.65 & - & - \\
& & RS & - & - & 0.8579 & 2.14 \\
\hline

\multirow{5}{*}{\tabincell{c}{\textbf{Student} \\ \textbf{Net}}} & LA & Both & 0.9067 & 3.87 & 0.7305 & 9.18 \\
& \tabincell{c}{ME \\ (LA+FA)} & Both & 0.9092 & 4.05 & 0.7340 & 8.42\\
\cline{2-7}
& \tabincell{c}{Ours \\ (MA+ME)} & Both & \textbf{0.9167} & \textbf{2.14} & \textbf{0.8969} & \textbf{8.14} \\
\toprule[1pt]
\end{tabular}
\end{table}

To validate the effectiveness of each component of our framework, we start from the baseline (single-organ segmentation using DeepLabV3+) and successively add our framework components. Specifically, our Model Adaptation (MA) stage includes two main components: the label refinement module (LRM) and the feature generalization module (FGM). Our Model Ensemble (ME) stage includes two components: label aggregation (LA) and feature aggregation (FA). As shown in Table ~\ref{tab_component}, the performance increases when adding each component and our proposed full method achieves the best results.

\subsubsection{Investigation on Pseudo label Refinement}
\label{Eff_LD}





\begin{table}[t]
\renewcommand\tabcolsep{5.5pt}
\renewcommand\arraystretch{1.1}
\centering
\scriptsize
\caption{Different strategies of pseudo-label refinement verified on the BTCV Dataset.}
\label{Table_Label_Denoise}
\begin{tabular}{ccccccc}
\toprule[1pt]

\multirow{2}{*}{\tabincell{c}{\textbf{Pseudo} \\ \textbf{Labeling}}}  && \multirow{2}{*}{\tabincell{c}{\textbf{Source} \\ \textbf{Domain}}} & \multicolumn{2}{c}{\textbf{Liver}} & \multicolumn{2}{c}{\textbf{Spleen}} \\
& & & DSC & ASD & DSC & ASD\\
\hline

\multirow{2}{*}{Prediction} & \multirow{2}{*}{} & LiTS & 0.9032 & 3.31 & - & - \\
& & RS & - & - & 0.7161 & 3.64 \\
\multirow{2}{*}{LRM} & \multirow{2}{*}{} & LiTS & \textbf{0.9064} & \textbf{3.65} & - & - \\
& & RS & - & - & \textbf{0.8579} & \textbf{2.14} \\

\toprule[1pt]
\end{tabular}
\end{table}

\begin{figure*}[t]
\centering  
\includegraphics[width=1\textwidth]{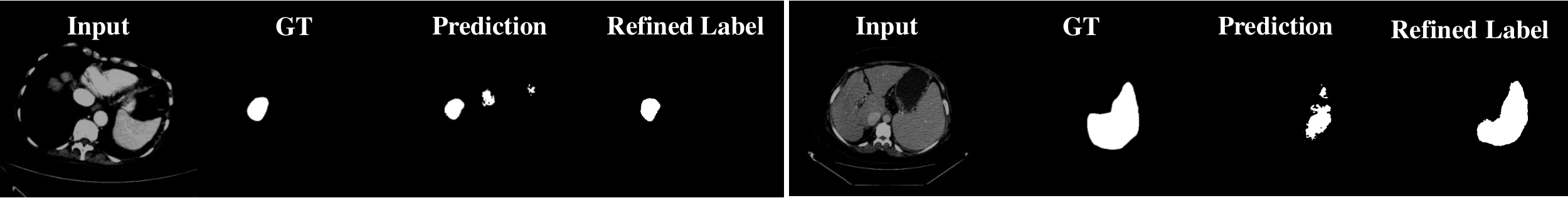}
\caption{Examples of refined pseudo-labels produced by the LRM.}
\label{Fig_Results_Label_Denoising}
\end{figure*}

To verify the effectiveness of the pseudo label refinement strategy used in the Model Adaptation stage, we first present some examples of refined labels generated by LRM in Fig.~\ref{Fig_Results_Label_Denoising}. It is shown that our module is effective in reducing pixel-wise noises especially when the predicted results are not confident. Besides, we also compare the LRM with the known pseudo-labeling method and the quantitative results are shown in Table ~\ref{Table_Label_Denoise}. Specifically, ``Prediction" means directly employing network predictions as pseudo labels.
From Table ~\ref{Table_Label_Denoise}, we can see that our LRM attains better performance and is effective in refining pseudo labels.

\subsubsection{Investigation on Ensemble Function}
\label{Eff_FF}

\begin{figure*}[t]
\centering  
\includegraphics[width=\textwidth]{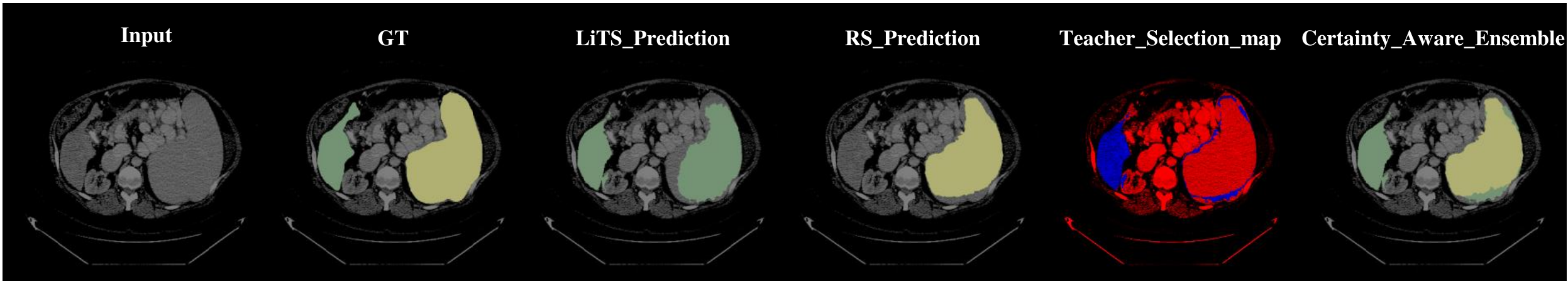}
\caption{Examples of ensemble results produced by Certainty-Aware Ensemble Function. The \textcolor[RGB]{34, 139, 34}{\textbf{green areas}} are liver segmentation results and the  \textcolor[RGB]{255, 193, 27}{\textbf{yellow areas}} are spleen segmentation results. The \textcolor[RGB]{178, 34, 34}{\textbf{red areas}} are the selection area of liver teacher model and the  \textcolor[RGB]{24, 116, 205}{\textbf{blue areas}} are the selection area of liver teacher model. }
\label{Fig_teacher_selection_map}
\end{figure*}

\begin{figure*}[t]
\centering  
\includegraphics[width=\textwidth]{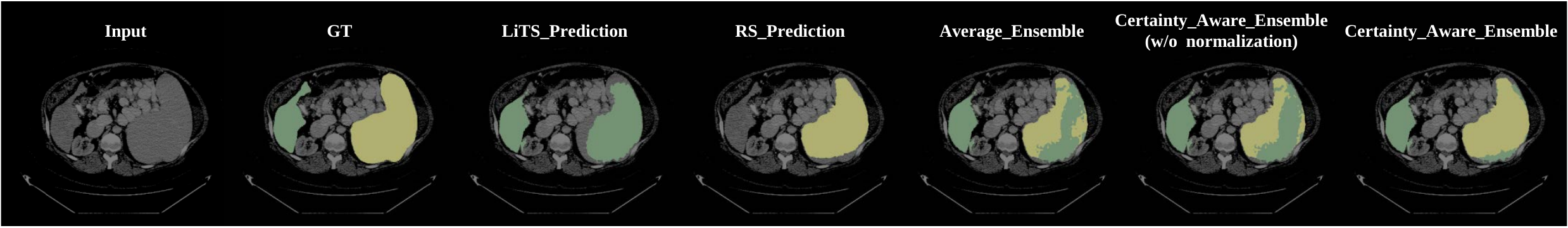}
\caption{Examples of ensemble results produced by Certainty-Aware Ensemble Function. The \textcolor[RGB]{34, 139, 34}{\textbf{green areas}} are liver segmentation results and the  \textcolor[RGB]{255, 193, 27}{\textbf{yellow areas}} are spleen segmentation results.}
\label{Fig_Results_Fusion_Function}
\end{figure*}

\begin{table}[t]
\renewcommand\tabcolsep{7pt}
\renewcommand\arraystretch{1.1}
\centering
\scriptsize
\caption{Different Ensemble Function verified on the BTCV Dataset.}
\label{Table_Results_Fusion_Function}

\begin{tabular}{ccccc}
\toprule[1pt]

 \multirow{2}{*}{\textbf{Ensemble Function}} &  \multicolumn{2}{c}{\textbf{Liver}} & \multicolumn{2}{c}{\textbf{Spleen}} \\
 & DSC & ASD & DSC & ASD\\
\hline

Average Ensemble & 0.9048 & 2.37 & 0.8419 & 1.90 \\
\tabincell{c}{Certainty-Aware Ensemble \\ (w/o normalization)} & 0.9136 & 3.07 & 0.8575 & 2.22 \\
Certainty-Aware Ensemble & \textbf{0.9167} & \textbf{2.14} & \textbf{0.8969} & \textbf{1.46} \\

\toprule[1pt]
\end{tabular}
\end{table}

To demonstrate the rationality of the ensemble function used in the Model Ensemble stage, we first provide the visual results of our ensemble function in Fig.~\ref{Fig_teacher_selection_map}. It is observed that our ensemble function resolves conflicts between multiple teacher nets and correctly segments the organ. We also replace our ensemble function with the average ensemble (directly average the softmax output of multiple teacher models) and certainty-aware ensemble without entropy normalization (w/o Eq.~\ref{Equ_Uniform}). The comparison results are displayed in Table ~\ref{Table_Results_Fusion_Function} and Fig.~\ref{Fig_Results_Fusion_Function}. By comparison, our proposed certainty-aware ensemble with entropy normalization exhibits the best performance.

\subsubsection{Investigation on Heterogeneous Teachers}
\label{stability}

\begin{table}[t]
\renewcommand\tabcolsep{3pt}
\renewcommand\arraystretch{1.1}
\centering
\scriptsize
\caption{Quantitative analysis and comparison of different network backbones verified on the BTCV Dataset. U is short for Unet backbone, and D is short for DeeplabV3+ Backbone. U+D represents combining a teacher net with the Unet backbone and a teacher net with the DeeplabV3+ Backbone.}
\label{tab_stability}
\begin{tabular}{cccccccc}
\toprule[1pt]

\multicolumn{2}{c}{\multirow{2}{*}{\textbf{Method}}} & \multirow{2}{*}{\textbf{Backbone}} & \multirow{2}{*}{\tabincell{c}{\textbf{Source} \\ \textbf{Domain}}} & \multicolumn{2}{c}{\textbf{Liver}} & \multicolumn{2}{c}{\textbf{Spleen}} \\
& & & & DSC & ASD & DSC & ASD\\
\hline

\multirow{4}{*}{\tabincell{c}{\textbf{Teacher} \\ \textbf{Net}}} & \multirow{4}{*}{\tabincell{c}{Baseline\\w/o \\domain\\ adaptation}} & \multirow{2}{*}{DeeplabV3+} & LiTS & 0.8969 & 4.20 & - & - \\
& & & RS & - & - & 0.6973 & 10.11 \\
& & \multirow{2}{*}{UNet} & LiTS & 0.9067 & 3.58  & - & - \\
& & & RS & - & - & 0.7438 & 7.96 \\
\hline

\multirow{4}{*}{\tabincell{c}{\textbf{Adapted} \\ \textbf{Teacher} \\ \textbf{Net}}} & \multirow{4}{*}{MA} & \multirow{2}{*}{DeeplabV3+} & LiTS & 0.9064 & 3.65 & - & - \\
& & & RS & - & - & 0.8579 & 2.14 \\
& & \multirow{2}{*}{UNet} & LiTS & 0.9111 & 2.95 & - & - \\
& & & RS & - & - & 0.7675 & 6.78 \\
\hline

\multirow{3}{*}{\tabincell{c}{\textbf{Student} \\ \textbf{Net}}} & \multirow{3}{*}{Ours} & D+D & Both & 0.9126 & 2.14 & 0.8969 & 1.46 \\
& & U+U & Both & 0.9156 & 2.58 & 0.7781 & 7.21 \\
& & U+D & Both & 0.9173 & 2.07 & 0.8714 & 2.98 \\
\toprule[1pt]

\end{tabular}
\end{table}

\begin{table*}[t]
\renewcommand\tabcolsep{3.4pt}
\renewcommand\arraystretch{1.1}
\centering
\scriptsize
\caption{Quantitative analysis of introducing a new single-organ model verified on the BTCV dataset. ``Ours (two-to-one)" represents combining two source models into one target model while ``Ours (three-to-one)" represents combining three source models into one target model.}
\label{tab_expansion}
\begin{tabular}{cccccccc}
\toprule[1pt]

\multirow{2}{*}{\textbf{Method}} & \multirow{2}{*}{\tabincell{c}{\textbf{Source} \\ \textbf{Domain}}}  & \multicolumn{2}{c}{\textbf{Liver}} & \multicolumn{2}{c}{\textbf{Spleen}} & \multicolumn{2}{c}{\textbf{Kidney}} \\
& & DSC & ASD & DSC & ASD & DSC & ASD\\ 
\hline

\textbf{Baseline (w/o adaptation)} & KiTS & - & - & - & - & 0.8497 & 2.46\\

\textbf{MA (ours)} & KiTS & - & - & - & - & 0.8829 & 1.34 \\
\hline

\textbf{Ours (two-to-one)} & LiTS+RS & 0.9126 & 2.14 & 0.8969 & 1.46 & - & - \\
\textbf{Ours (three-to-one)} & LiTS+RS+KiTS & \textbf{0.9182} & \textbf{2.07} & \textbf{0.8984} & \textbf{1.51} & \textbf{0.9282} & \textbf{1.27}\\
\toprule[1pt]

\end{tabular}
\end{table*}

We further consider combining knowledge from teacher models with different network architectures (backbones). Specifically, we replace the backbone from DeepLabV3+~\cite{chen2017deeplab} to Unet~\cite{ronneberger2015u} and the results are shown in Table ~\ref{tab_stability}. The experimental results demonstrate that our proposed method is robust to different backbones, implying the stability and generalization of our method. Meanwhile, the results of multi-source (``D+D", ``U+U", and ``U+D") also show that our method can flexibly merge knowledge from teacher models with different network structures, which suggests that our framework has good application prospects.

\subsubsection{Extension to New Single-Organ Segmentation Model}
\label{expansion}

To further verify the applicability and adaptability of our proposed framework, we extend our model to a more complex scenario in which a new single-organ segmentation model (kidney) is introduced. The new single-organ segmentation model is trained on the Kidney tumor segmentation challenge (KiTS) dataset~\cite{heller2021state}, which contains 210 CT scans (size range: $[29 \sim 1059]\times[512]\times\{512,796\}$ voxels) with annotation for kidney and kidney tumors. The in-plane spacing of CT slices varies from 0.44mm to 1.04mm, and the slice thickness varies from 0.5mm to 5.0mm. In the new Multi-Model Adaptation scenario, we aim to aggregate three single-organ models into one multi-organ model. The results are shown in Tab.~\ref{tab_expansion}, in which ``Ours (two-to-one)" represents combining two source models into one target model while ``Ours (three-to-one)" represents combining three source models into one target model. Experimental results demonstrate that our framework is still effective when introducing a new dataset. Besides, we also show that our framework is flexible enough to be extended to combine any number of source models.

\section{Discussion and Conclusions}
In this paper, we present a novel approach for multi-organ segmentation that addresses the challenge of limited annotated data. Specifically, we propose a Multi-Model Adaptation framework that leverages off-the-shelf single-organ segmentation models to learn a multi-organ segmentation model. The framework comprises two stages: Model Adaptation and Model Ensemble. In the Model Adaptation stage, we fine-tune each single-organ segmentation model to improve its generalization on unseen target datasets. In the Model Ensemble stage, we aggregate the knowledge from multiple adapted models to obtain a robust multi-organ segmentation model.

We conducted extensive experiments on four abdominal image datasets, which demonstrated the feasibility and effectiveness of our approach. Despite advancements achieved, our method still falls short in terms of segmentation accuracy compared with supervised segmentation methods, which may limit its clinical applicability. Future research could focus on improving the segmentation performance by incorporating more knowledge about the organs, such as using shape adversarial prior\cite{yin2022transfgu} and active learning in such setting\cite{su2020active}, and exploring the possibility of combining partial annotated models not limited to single-organ models.

\section*{Acknowledgment}
This work was supported in part by the Major Scientific Research Project of Zhejiang Lab under Grant 2020ND8AD01, the Jiangsu Funding Program for Excellent Postdoctoral Talent.


\bibliography{ref}

\end{document}